\documentclass[12pt,a4paper]{article}
\pdfoutput=1

\usepackage{setspace}
\usepackage{caption}
\usepackage{longtable,multicol}
\usepackage{soul}
\usepackage[]{graphicx}
\usepackage{lineno}
\usepackage[FIGTOPCAP]{subfigure}
\usepackage{amssymb}
\usepackage{epsfig,amsfonts,amsmath}\usepackage{wrapfig}
\usepackage{lipsum}
\usepackage{gensymb}
\usepackage{relsize}
\usepackage{exscale}
\usepackage{dcolumn}   
\usepackage{bm}   
\usepackage[rightcaption]{sidecap}
\usepackage[top=3cm, bottom=3cm, left=2cm, right=2.5cm]{geometry}
\usepackage{hyperref} 
\usepackage{color,soul}
\usepackage{parskip}
\newcommand{\vect}[1]{\boldsymbol{\mathit{#1}}}

\newcommand{\tens}[1]{\mathbf{#1}}

\begin{document}

\title{Reduced-PINN: An Integration-Based Physics-Informed Neural Networks for Stiff ODEs}
\author{Pouyan Nasiri\textsuperscript{a}, Roozbeh Dargazany\textsuperscript{*a} \\
\small \emph{\textsuperscript{a}Department of Civil and Environmental Engineering, Michigan State University, USA}
\\
\small \emph{\textsuperscript{*}corresponding author:e-mail:roozbeh@msu.edu} \normalsize}

\maketitle

\singlespacing
\begin{abstract}
Physics-informed neural networks (PINNs) have recently received much attention due to their capabilities in solving both forward and inverse problems. For training a deep neural network associated with a PINN, one typically constructs a total loss function using a weighted sum of different loss terms and then tries to minimize that. This approach often becomes problematic for solving stiff equations since it cannot consider adaptive increments. Many studies reported the poor performance of the PINN and its challenges in simulating stiff chemical active issues with administering conditions of stiff ordinary differential conditions (ODEs). Studies show that stiffness is the primary cause of the failure of the PINN in simulating stiff kinetic systems.  

Here, we address this issue by proposing a reduced weak-form of the loss function, which led to a new PINN architecture, further named as Reduced-PINN, that utilizes a reduced-order integration method to enable the PINN to solve stiff chemical kinetics. The proposed Reduced-PINN can be applied to various reaction-diffusion systems involving stiff dynamics.
To this end, we transform initial value problems (IVPs) to their equivalent integral forms and solve the resulting integral equations using physics-informed neural networks. In our derived integral-based optimization process, there is only one term without explicitly incorporating loss terms associated with ordinary differential equation (ODE) and initial conditions (ICs). To illustrate the capabilities of Reduced-PINN, we used it to simulate multiple stiff/mild second-order ODEs. We show that Reduced-PINN captures the solution accurately for a stiff scalar ODE. We also validated the Reduced-PINN against a stiff system of linear ODEs. In the last step, we used the Reduced-PINN to simulate the ROBER problem, a significant challenge made by a system of nonlinear ODEs. Our numerical validations show the capability of Reduced-PINN to represent stiff ODEs at a significantly reduced computational cost.

\end{abstract}

\section{Introduction}

Ordinary and partial differential equations (ODEs and PDEs) are main drivers of the simulation cost in many large-scale simulations, ranging from physical events, chemical reactions, and economic models. In order to solve them, there are a variety of classical methods, e.g., finite element and finite difference methods \cite{ref1}. These numerical procedures depend on the discretization of the spatiotemporal domain and then solve the resulting linear or nonlinear algebraic equations. As a result, the accuracy of classical numerical approaches relies directly on the element size or the grid size of the discrete version of governing equations. Physics-informed neural networks (PINNs) proposed by Raissi et al. \cite{ref2} is an alternative for solving ODEs or PDEs. This method overcomes the need to discretize the governing equations. Instead, we need to find the parameters (weights and biases) of a deep neural network through training process. Because of the universal approximation theorem, a deep neural network can approximate a wide range of functions with an arbitrary precision.

PINNs have been successfully applied to many problems from numerous disciplines including solid mechanics \cite{ref3,ref4,ref5,ref6,ref7} and fluid mechanics \cite{ref8,ref9}. In addition, other versions of physics-informed neural networks such as conservative PINNs \cite{ref10} and variational PINNs \cite{ref11,ref12} allow more flexibility in comparison with regular PINNs. The total loss function of PINNs involves some terms related to differential equations, initial conditions, and boundary conditions. 
ODEs are widely used to describe the evolution of the species concentrations in chemical kinetics. Predicting stiff chemical systems is fundamental in modeling most real-time chemical reactions ranging from energy storage to material aging and biomedical design. A kinetic equation is called stiff when some involved species evolve slowly while others rapidly change in a short time frame. This forces the numerical method to take small steps over a long range to obtain satisfactory results. Thus, stiffness highly affects the computational efficacy of the solution.
 
It is known that regular PINN fails to represent stiff chemical systems \cite{ref13} due to the numerical stiffness of the involved ODEs. Weiqi et al. \cite{ref13} solved stiff chemical ODEs by imposing steady-state assumptions on some state variables. As a result, they found the solution to a non-stiff/mild system of ODEs instead of the original stiff ODEs. In a recent study, De Florio et al. \cite{ref14} solved the stiff system of nonlinear ODEs without transformation to non-stiff/mild equations. However, their approach is based on a shallow NN and requires solving a nonlinear system of equations for a general system of ODEs. 

To enable PINN for stiff systems, Wang et al. \cite{ref15} proposed a learning rate annealing procedure to balance the interaction among loss function terms. The weights of terms in the total loss function can be adaptively modified after each epoch or a few epochs.

In this paper, we develop Reduced-PINN, the first weak-form of PINN framework developed for simulating the stiff ODEs based on derivation of the weak form of the involved ODEs. In particular, our main objectives in this paper are to derive
\begin{enumerate}
	\item \textit{Weak-form formulation}: we transform a regular PINN for a scalar ODE with multiple loss terms to an integral-based PINN with a single loss term. To this end, an initial value problem is converted into an integral equation and solved by physics-informed neural networks. The suggested method ensures that we are treating the same problem with only one loss term.
	\item \textit{Long-time simulation of stiff systems}: we will apply Reduced-PINN sequentially for long-term simulations. If we divide a long time interval into smaller time steps, it provides us with two main advantages: one is faster training for each Reduced-PINN and the other is to approximate the involved integrals with explicit formulas.
\end{enumerate}

 In this paper, we developed and validated a new engine, Reduced-PINN, to simulate stiff systems using weak-form of loss function. The transformation process to convert initial value problems (IVPs) into weak-form integral equations is given in section 2. We discuss classical PINN architecture and how to implement Reduced-PINN in sections 3 and 4. To validate our engine, the performance of Reduced-PINN is studied in multiple mild/stiff problems in section 5. Finally, conclusions and the outlook for future work will be presented in section 6.

\section{\textbf{PROBLEM SETUP}}
\paragraph {Stiff Chemical Systems} Multiple techniques were developed to derive the numerical solution of the above ODE, such as implicit
Runge-Kutta time-stepping schemes \cite{ref16}, although they are usually constrained due to the
computational cost of solving Eq. (\ref{eq1}).
In chemical kinetics, most ODE systems that represent the kinetic models are stiff, and thus require ultra-fine time-steps for simulation which makes time-domain integration computationally expensive. 
Regardless of the nature of reactions, it is often difficult to provide a generalized description for the stiffness of a chemical system, and usually we identify stiffness of a system by the distinct time scales of their species. Fast-reacting species operate on short time scale, whereas others may react slower and have orders of magnitude more extended periods. 
Accordingly, one can consider an ODE system to be stiff if the computational cost of classical non-stiff ODE solver becomes much higher than implicit ODE integrators.

\paragraph {ODE of Stiff Systems} A first-order non-homogeneous chemical reaction system can be modeled using the following ODE
\begin{align}
\label{eq1}
\vect{u}^{(1)}(x) + \tens{A}(x)\vect{u}(x)& = \vect{f}(x), \qquad \forall x \in \left[a,b\right] \nonumber\\ 
\vect{u}(x)=\left[u_1(x),..,u_N(x)\right]^T, \qquad {u}_i(a)&={\mu}_i^{<0>}, \qquad \vect{u}^{(n)}(x) =\frac{d^n\vect{u}}{dx^n}.
\end{align}
where the superscript $\bullet^{<j>}$ represents the parameter index, $\bullet^{(j)}$ the order of differentiation, and the subscript $\bullet_j$ represents parameters associated to $j$-th species. Hereafter, we use the following formatting; $\tens{A}$ to represent tensors, $\vect{a}$ to represent vectors, and ${a}$ to represent scalar parameters.

\paragraph {Weak-Form Derivation} Here, we elaborate the process of converting a general ODE problem into an integral equation. The procedure is derived for each individual species $u_i(x)$. 
For simplicity, here we assume $\vect{u}(x)$ to be a single component vector which represents the concentration of one specie at the reaction time $x\in \left[a,b\right]$ with initial value to be $\mu^{<0>}:=\mu_1^{<0>}$. However, the formulation can be further generalized to note that for multi-component vector $\vect{u}(x)$, the weak-form derivation will be implemented for each component $u_i(x)$. 
Let us assume a kinetic reaction with the following ODE 
\begin{equation}
\label{eq2}
{u}^{(n)}(x) + \lambda_{1}(x) u^{(n-1)}(x) +\dots + \lambda_n(x){u}(x)= f(x) \
\end{equation}
with initial conditions
\begin{equation}
\label{eq3}
u(a) = \mu^{<0>}, u^{(1)}(a) = \mu^{<1>}, \dots, u^{(n-1)}(a) = \mu^{<n-1>}
\end{equation}
where $\lambda_i(x) (i=1,\dots,n) $ and $ f(x) $ are continuous and given for $a\le x \le b$. Also, the $n$ in Eq. (\ref{eq2}) represents the order of ordinary differential equation. We need the following identity to convert the above IVP into an integral equation
\begin{equation}
\label{eq4}
\underbrace{\int_a^x\int_a^{x_1}\dots\int_a^{x_{m-1}}}_{m \ \mathrm{times}} v(x_{m}) dx_{m}dx_{m-1}\dots dx_1= \int_{a}^{x}\frac{(x-t)^{m-1}}{(m-1)!} v(t)dt
\end{equation}
To prove this identity, we adopt the approach given in \cite{ref17}. Let $I_{m} v(x)$ be defined as
\begin{equation}
\label{eq5}
I_{m} v(x) = \int_{a}^{x}\frac{(x-t)^{m-1}}{(m-1)!} v(t)dt
\end{equation}
By using the Leibniz rule and differentiating both sides of Eq. (\ref{eq5}), we have 
\begin{equation}
\label{eq6}
\frac{d}{dx}(I_{m} v(x)) = \frac{(x-x)^{m-1}}{(m-1)!} v(x) + \int_{a}^{x}\frac{(x-t)^{m-2}}{(m-2)!} v(t)dt
= I_{m-1} v(x)
\end{equation}
In addition, we have $I_{m} v(a)= 0 $ for any $m\geq1$. If we integrate both sides of Eq. (\ref{eq6}), we arrive at

\begin{equation}
\label{eq7}
I_{m} v(x)= I_{m} v(a) + \int_{a}^{x} I_{m-1} v(\eta) d\eta = \int_{a}^{x} I_{m-1} v(\eta) d\eta 
\end{equation}
If we use the above equation successively, we derive the identity presented in Eq. (\ref{eq4}). Now, let us define $u^{(n)}(x) = v(x)$, by integrating this expression one time, we obtain
\begin{equation}
\label{eq8}
u^{(n-1)}(x) = \int_{a}^{x} v(t)dt + u^{(n-1)}(a) = \int_{a}^{x} v(t)dt + \mu^{<n-1>}
\end{equation}
Integrating one more time, we have
\begin{equation}
\label{eq9}
u^{(n-2)}(x) = \int_{a}^{x}\int_{a}^{x_1} v(x_2) dx_2 dx_1+ \mu^{<n-1>} (x-a) + \mu^{<n-2>}
\end{equation}
Continuing this process, the following general expression for $(k=0,\dots,n-1)$ is derived
\begin{equation}
\label{eq10}
u^{(k)}(x) = \underbrace{\int_a^x\int_a^{x_1}\dots\int_a^{x_{n-k-1}}}_{n-k \ \mathrm{times}} v(x_{n-k}) dx_{n-k}dx_{n-k-1}\dots dx_1 + \sum_{i=0}^{n-k-1}\mu^{<n-i-1>} \tfrac{(x-a)^{n-k-i-1}}{(n-k-i-1)!}
\end{equation}
 where we define $u^{(0)}(x) = u(x)$. The first expression of Eq. (\ref{eq10}) can be simplified further by the identity given in Eq. (\ref{eq4}). As a result, we have
\begin{equation}
\label{eq11}
u^{(k)}(x) = \int_{a}^{x}\frac{(x-t)^{n-k-1}}{(n-k-1)!} v(t)dt+ \sum_{i=0}^{n-k-1}\mu^{<n-i-1>} \tfrac{(x-a)^{n-k-i-1}}{(n-k-i-1)!}
\end{equation}
The above relationship holds for an arbitrary derivative of $u(x)$. In a compact form, Eq. (\ref{eq2}) can be written as 
\begin{equation}
\label{eq12}
u^{(n)}(x) + \sum_{j=0}^{n-1}\lambda_{n-j}(x) u^{(j)}(x) = f(x)
\end{equation}
If we substitute Eq. (\ref{eq11}) into the above expression for each derivative of $u(x)$ and use $u^{(n)}(x) = v(x)$, we finally derive the following integral equation
\begin{equation}
\label{eq13}
v(x) + \int_{a}^{x} \psi(x,t)v(t) dt= g(x)
\end{equation}
where 
\begin{equation}
\label{eq14}
\psi(x,t) = \sum_{j=0}^{n-1}\lambda_{n-j}(x)\frac{(x-t)^{n-j-1}}{(n-j-1)!}
\end{equation}
and
\begin{equation}
\label{eq15}
g(x) = f(x) - \sum_{j=0}^{n-1}\sum_{i=0}^{n-j-1}\mu^{<n-i-1>}\lambda_{n-j}(x)\frac{(x-a)^{n-j-i-1}}{(n-j-i-1)!}
\end{equation}
The residual associated with the integral equation is written as
\begin{equation}
\label{eq16}
R(x) = v(x) + \int_{a}^{x} \psi(x,t)v(t) dt - g(x)
\end{equation}
\
The interested reader can refer to \cite{ref17} for a detailed discussion on integral equations.

\section{Classical PINN Architecture}

\begin{SCfigure}[40][tbh]
\centering
{\includegraphics[width =0.5\linewidth]{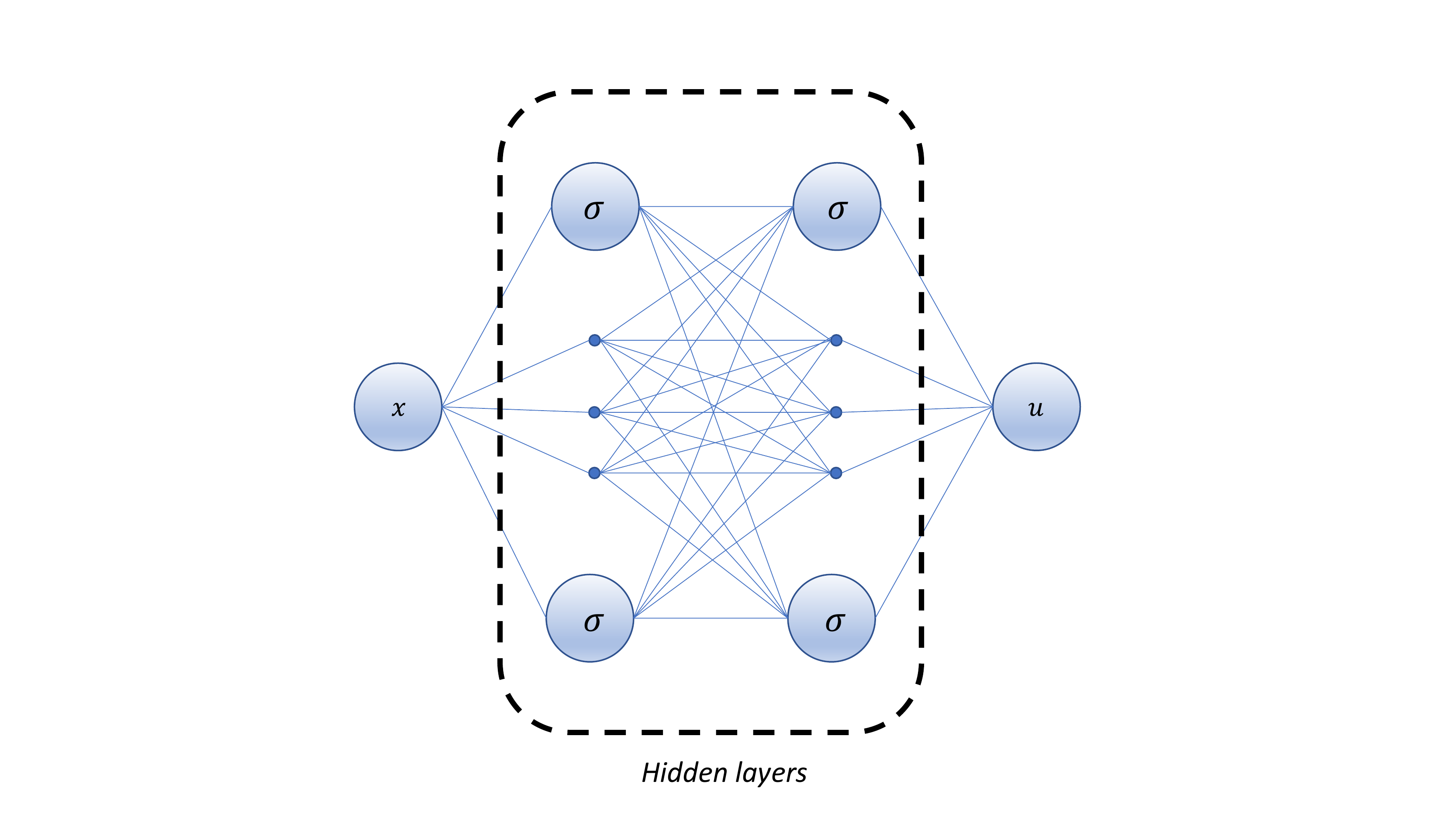}}
\caption{Schematic representation of a deep neural network trained with Reduced-PINN method.
}
\label{fig1}
\end{SCfigure}

\
The success of physics-informed neural networks is essentially based on the ability of deep neural networks to approximate a vast number of functions appearing in practice. In an initial value problem, a neural network can be set up to represent $\vect{u}(x)$ with $x$ as input, and $u$ as the output as 
\begin{equation}
\label{eq17}
\vect{u}(x) = l(\Phi^{<h>}(\Phi^{<h-1>}\dots(\Phi^{<0>}(x)))), \qquad \Phi^{<i>}(\hat{x}) = \sigma(W^{<i>} \cdot \hat{x} + b^{<i>})
\end{equation}
where $\theta=\left[\left[W^{i}\right],\left[b^i\right]\right]$ is the set of weights $W^{i}$ and biases $b^{i}$, $\sigma$ a nonlinear activation function, and $l$ a linear amplification function.
 Note that $u=\left[u_1,..,u_N\right]^T$ represents the vector of species concentrations, and $N$ is the number of acting chemical species in the ODE.
Fig. (\ref{fig1}) shows a schematic representation of a fully connected feed-forward neural network with two hidden layers and arbitrary units in each layer. As can be seen from this figure, there is only one output to this neural network because we have used separate neural networks corresponding to each state variable, and that is only our preference in this study. It should be noted that any type of feed-forward neural network including shallow networks can be used for implementation. 

In comparison to recent studies, the proposed Reduced-PINN has two main advantages over Stiff-PINN methods \cite{ref13,ref14}; (1) Our method uses a deep NN as well as a shallow NN, and be trained by back-propagation for any nonlinear system of ODEs. (2) Our method solves the original equations without any transformation to non-stiff/mild ones.

\paragraph{Loss function of classical PINN } is derived by implementing automatic differentiation \cite{ref18} as additional network to NN representing $\vect{u}(x)$. Accordingly, by substituting derivatives of $\vect{u}(x)$ in the ODE by automatic differentiated NN, one derives the ODE residual term $\mathcal{L}_{ODE} (\theta)$, and initial condition residual form $ \mathcal{L}_{IC_{i}}(\theta)$ as given below and furthers calculates the total loss function $\mathcal{L}(\theta)$ of a PINN as 

\begin{equation}
\label{eq18}
\mathcal{L}(\theta) = \mathcal{L}_{ODE} (\theta) + \sum_{i}^{} \alpha_{i} \mathcal{L}_{IC_{i}}(\theta)
\end{equation}
where $\alpha_{i}$ are fixed or adaptively tuned parameters. Now, by minimizing the total loss with respect
to the weights and biases, PINN solution is derived.

 \section{Reduced-PINN Implementation} 
\paragraph{Loss function of proposed Reduced-PINN } In the proposed integral-based PINN, our total loss function $\mathcal{L}(\theta)$ involves only one term since we have coupled the ODE and IVP equations into one integral equation. The residual defined in Eq. (\ref{eq16}) becomes a function of $\theta$ and $x$ after approximating $v(x)$ by a deep neural network with the parameters $\theta$. In this study, we consider the following loss function

\begin{equation}
\label{eq19}
\mathcal{L}(\theta) = \frac{1}{n}\sum_{i=1}^{n}R(x_{i};\theta)^{2}
\end{equation}
where $R(x;\theta)$ is the residual of the integral equation defined in Eq. (\ref{eq16}), and $n$ denotes the number of collocation points in time domain $x\in\left[a,b\right]$. In this paper, for each state variable, we have considered a separate neural network. Therefore, for a system of ODEs containing $M$ equations, we have the following total loss
\begin{equation}
\label{eq20}
\mathcal{L}(\Theta) = \sum_{i=1}^{M}\mathcal{L}(\theta_i)
\end{equation}
where $\Theta$ stands for the neural network parameters of all state variables, and $\mathcal{L}(\theta_i)$ is the loss corresponding to any scalar ODE of our system of ODEs.

 \paragraph{Numerical Integration} Using TensorFlow \cite{ref19} for numerical implementation, the integrals are calculated based on the Newton–Cotes rules with the following general form
\begin{equation}
\label{eq21}
\int_{a}^{b}h(x) dx \approx \sum_{i=0}^{p}\omega_i h(x_i)
\end{equation}
where $\omega_i$ are weights, and $x_i$ are the equi-spaced timesteps within the time domain $[a, b]$.
 Often, a for loop is utilized for a general integration procedure, but this method for a Reduced-PINN takes a long time in terms of computational cost. A feasible approach to speed up the process of integration is to combine the Newton–Cotes rules (e.g., the trapezoidal rule) with the current capabilities of TensorFlow. We have used this technique in our numerical integration wherever possible. 

Using trapezoidal rule, numerical integration in Eq. (\ref{eq21}) is given as
\begin{equation}
\label{eq22}
\int_{a}^{b}h(x) dx \approx (\frac{\Delta x}{2} ) (h(x_0)+ 2 \sum_{i=1}^{p-1}h(x_i) + h(x_p))
\end{equation}
where $\Delta x = (\frac{b-a}{p})$, and $(x_i)_{i=0}^{p}$ is assumed to be a uniform partition of the domain. Now, for the summation in the middle of the above expression, we could use the reduce mean command from TensorFlow to evaluate the integral much faster than the usual for loop. In addition, for the sequential implementation of Reduced-PINN, we also have used the following Simpson's rule
\begin{equation}
\label{eq23}
\int_{a}^{b}h(x) dx \approx (\frac{b-a}{6}) (h(a)+ 4 h(\frac{a+b}{4}) + h(b))
\end{equation}
We could implement the above formula into our code directly, which gives an accurate approximation to the exact value of the integral when the difference between the end points of the interval of integration (i.e., $b-a$) is relatively small.

\begin{figure}[tbh]
\begin{center}
\includegraphics[width = 8cm]{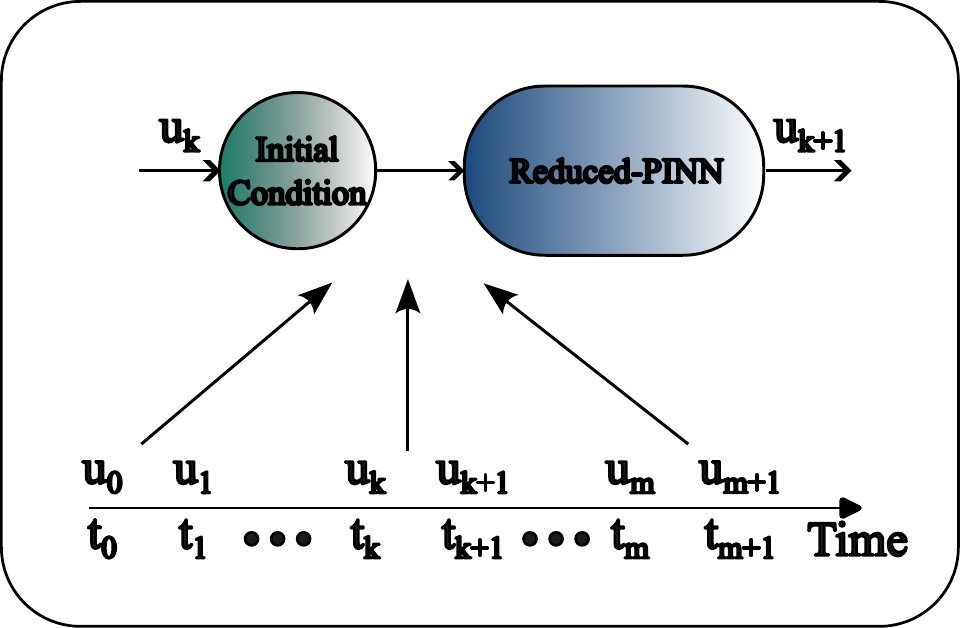}
\caption{Schematic representation of Sequential Reduced-PINN}
\label{main3}
\end{center}
\end{figure}

 \paragraph{System of ODEs} For completeness, we consider a general system of nonlinear ODEs as follows
 \begin{equation}
 \label{eq24}
\vect{u}^{(1)}(x)=\vect{q}(\vect{u}(x),x), \quad \Rightarrow \quad
 \begin{cases}
 u_{1}^{(1)}(x) = q_1(u_1,u_2,\dots,u_N, x)\\
 \vdots\\
 u_{N}^{(1)}(x) = q_N(u_1,u_2,\dots,u_N, x)
 \end{cases} 
 \end{equation}
with initial conditions $u_{i}(a) = \mu^{<0>}_i$ for $i=1,2,\dots,N$. We could use the following relations to transform the above equations into an appropriate framework for Reduced-PINN implementation
\begin{equation}
\label{eq25}
u_{i}^{(1)}(x)= v_{i}(x), i = 1,2,\dots,N.
\end{equation}
\begin{equation}
\label{eq26}
u_{i}(x) = \int_{a}^{x} v_{i}(t)dt + \mu^{<0>}_i, i = 1,2,\dots,N.
\end{equation}
Next, we approximate each unknown function $v_{i}(x)$ by a separate neural network, all with the same initial random values. After training, we will recover the Reduced-PINN solution to our IVP from Eq. (\ref{eq26}). Our numerical experiments show that, for practical cases, we should use Reduced-PINN sequentially. In other words, as shown in Fig. (\ref{main3}), the total time interval is divided into $m+1$ subintervals and the solution of the current Reduced-PINN at the end of its corresponding subinterval is given to the next Reduced-PINN as the initial condition. For the first Reduced-PINN, the initial conditions will be the same as the ones given in the problem definition. For optimization procedure, Adam optimizer will be used as a classical gradient-descent algorithm.

 \section{Numerical Results} 
 In this section, the performance of the proposed Reduced-PINN in different cases of mild and stiff ODEs will be benchmarked against solutions.
In the illustrated examples, classical PINN mostly fails due to the stiffness, and the performance of the Reduced-PINN in reducing stiffness and predicting the species concentrations will be provided. 
The first example is a mild ODE of second-order where we introduce and use sequential Reduced-PINN to solve it. The second case is a stiff scalar ODE of the first order. Next, we consider a stiff system of linear ODEs. Finally, we solve the ROBER problem. For training, an unknown function from integral equations is represented by a deep neural network. For all cases, our neural network has the following hyper-parameters: 4 hidden layers, 50 units per layer, ReLU activation function, and Adam optimizer. Afterward, we obtain the solution to our IVP by related integral formulas.
All examples addressed here was coded in Python and ran on Apple M1 chip, and 16 GB of RAM.

\subsection{Case Study 1: Mild ODE}
\begin{figure}[t]
\footnotesize{a)}{\includegraphics[width = 0.3\linewidth]{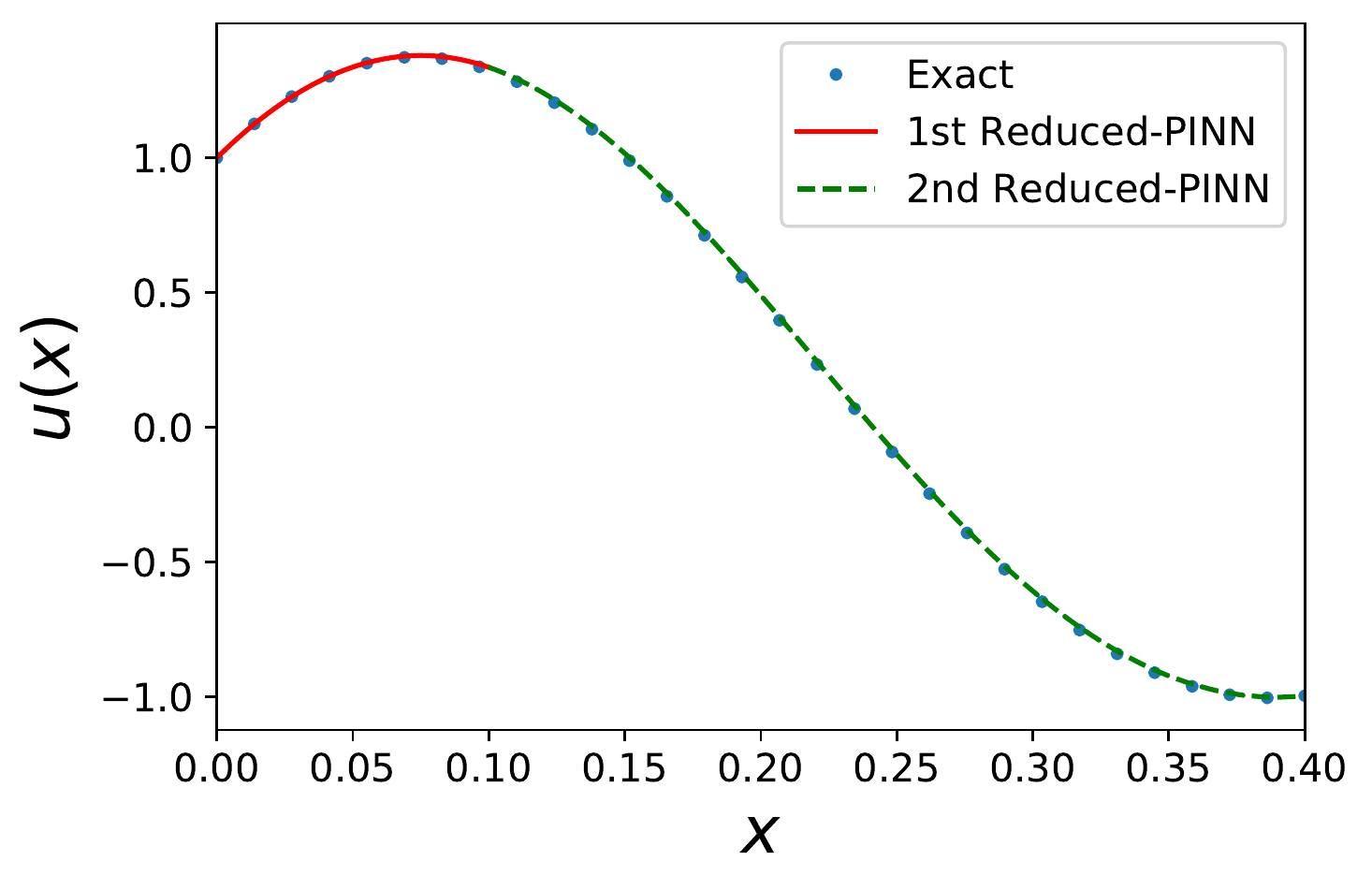}}
\footnotesize{b)}{\includegraphics[width = 0.3\linewidth]{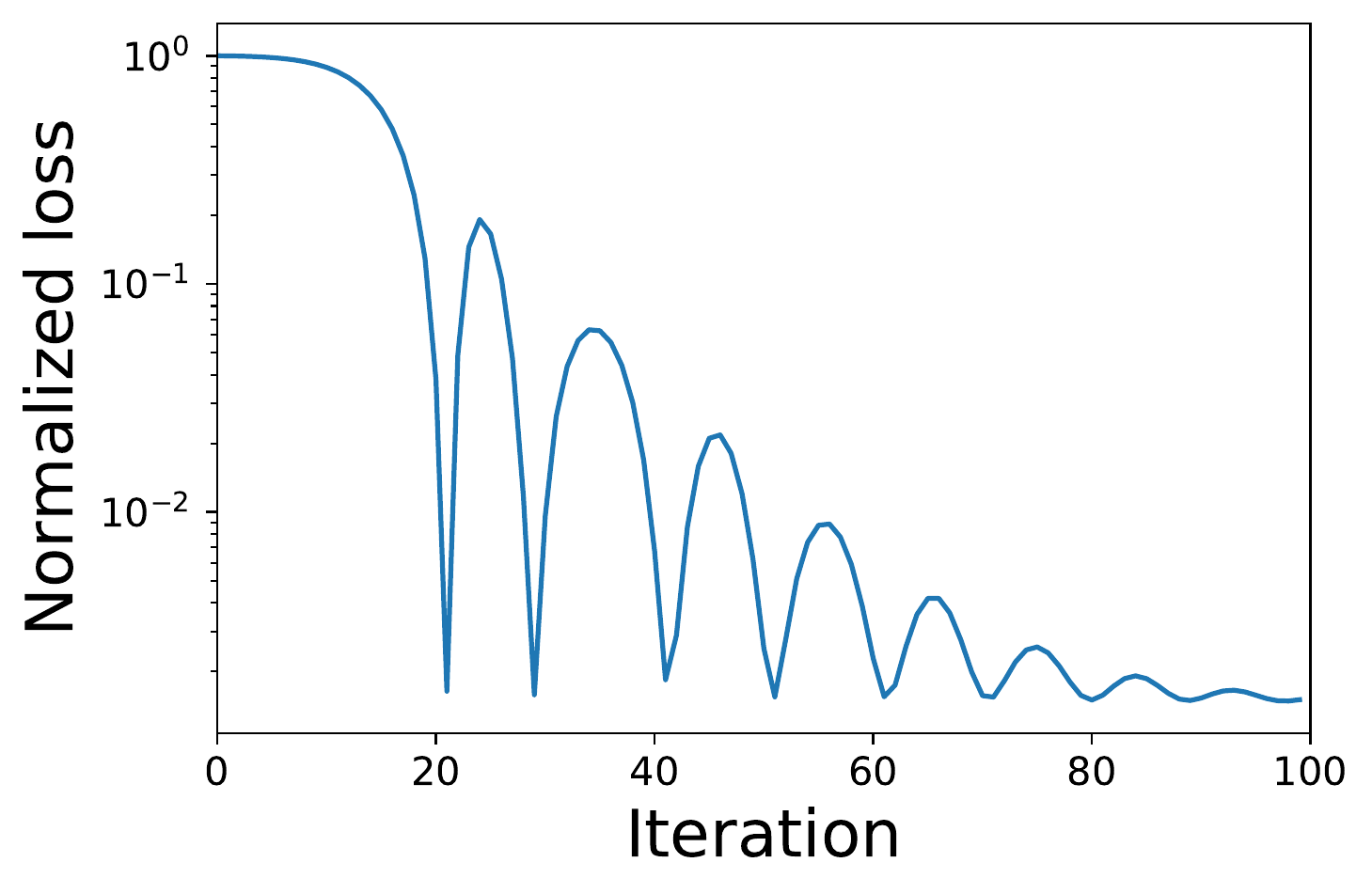}}
\footnotesize{c)}{\includegraphics[width = 0.3\linewidth]{convergence_pinn1.pdf}}
\caption{Case Study 1: Mild ODE example. a)  Exact solution vs. Reduced-PINN predictions, b) Convergence history of  the first Reduced-PINN, c) Convergence history of  the second Reduced-PINN.}
\label{fig24}
\end{figure}

We first employed Reduced-PINN to solve the following second order initial value problem as a mild ODE
\begin{equation}
\label{eq27}
u^{(2)}(x) - 2cu^{(1)}(x)+(c^2+d^2)u(x)=0, (0\le{x}\le0.4)
\end{equation}
\begin{equation}
\label{eq28}
\ u(0)=\mu^{<0>}, u^{(1)}(0)=\mu^{<1>}.
\end{equation}
where $c$ and $d$ are some chosen values and $\mu^{<0>}$ and $\mu^{<1>}$ are arbitrary initial conditions for use in the sequential Reduced-PINNs, i.e., the solution of current Reduced-PINN is transferred to the next Reduced-PINN as initial conditions. The exact solution of the above differential equation is written as 
\begin{equation}
\label{eq29}
u_{exact}(x) = \left(\frac{\mu^{<1>}-c\mu^{<0>}}{d}\right)e^{cx}\sin{dx}+\mu^{<0>}e^{cx}\cos{dx}
\end{equation}
and its equivalent integral form is written as
\begin{equation}
\label{eq30}
v(x) + \int_{0}^{x}[(c^2+d^2)(x-t)-2c]v(t)dt+\mu^{<1>}[(c^2+d^2)x-2c]+\mu^{<0>}(c^2+d^2)=0
\end{equation}
after training, the Reduced-PINN solution to our problem is derived from 
\begin{equation}
\label{eq31}
u(x) = \int_{0}^{x}(x-t)v(t)dt + \mu^{<1>}x+\mu^{<0>}
\end{equation}
As an example, we choose $c = -1 $, $d= 10$, $\mu^{<0>} = 1$, and $\mu^{<1>} = 10 $ and consider using two sequential Reduced-PINNs. The first Reduced-PINN was implemented on the interval $[0,0.1]$ and its predictions were given to the second Reduced-PINN, which is implemented on $[0,0.3]$. In Fig. (\ref{fig24}a), we benchmarked Reduced-PINN predictions against the exact solution,  which shows an excellent agreement. The history of normalized loss value functions in the first and second Reduced-PINNs is also demonstrated in Fig.\ref{fig24}b and Fig. \ref{fig24}c, respectively. To get an accurate prediction by sequential Reduced-PINNs, one has to ensure that each Reduced-PINN is optimised to yield its best solution since the solutions are built upon each other. 

For this problem, we have used two sequential Reduced-PINNs as we can capture the solution with only two Reduced-PINNs. However, we could use many more Reduced-PINNs if we have a large time span. To elaborate more, initially, we divide the whole time span into smaller time subintervals. In the first time subinterval, we solved the ODE by using the initial conditions given by the problem. Thus, we approximate the main variables at the end of the first subinterval, which become the new initial conditions for the next Reduced-PINN solver, and so on.

%
%
%
%
 
\subsection{Case Study 2: Stiff Single ODE}
\begin{figure}[t]
\footnotesize{a)}{\includegraphics[width = 0.46\linewidth]{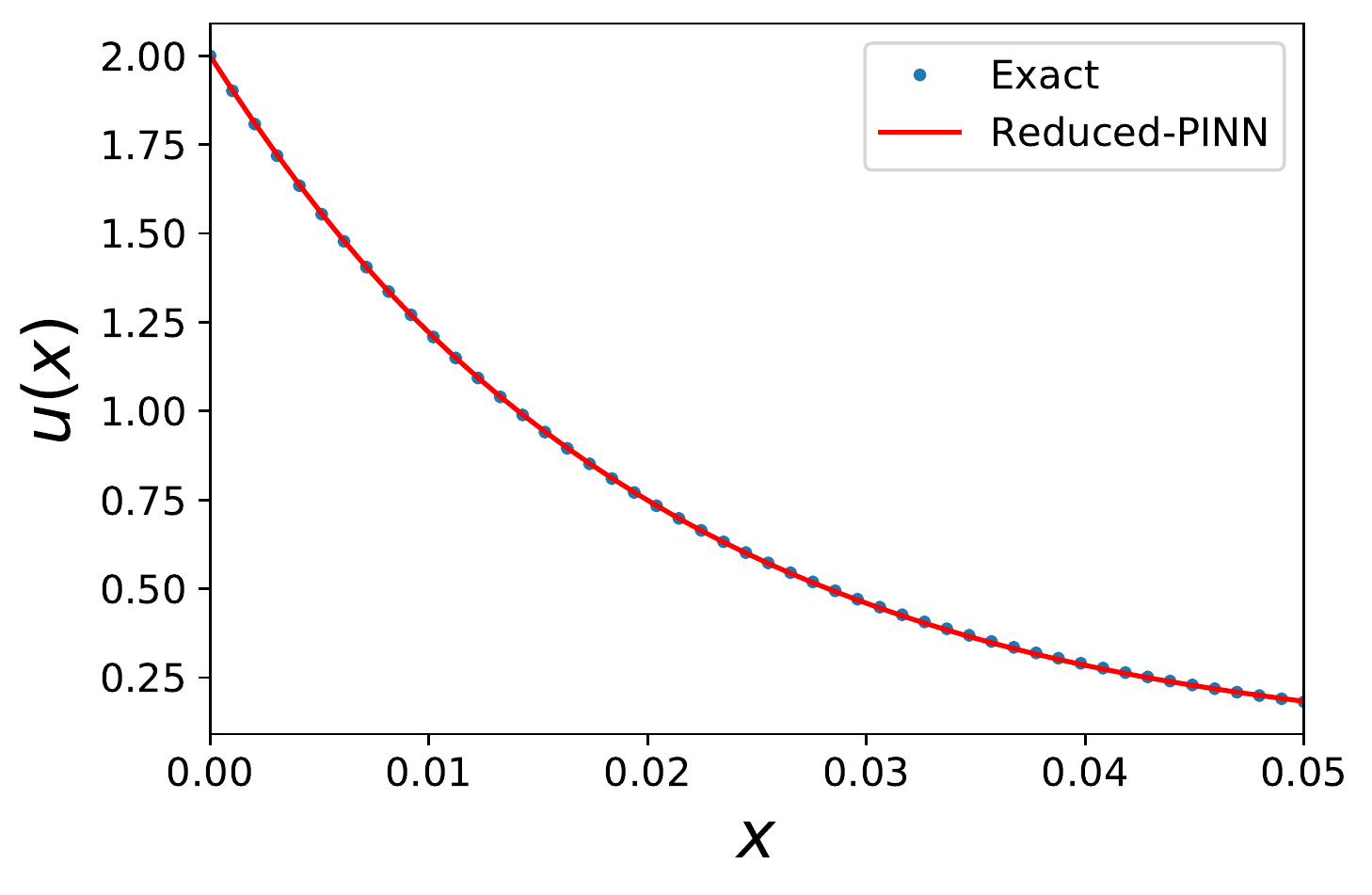}}
\footnotesize{b)}{\includegraphics[width = 0.46\linewidth]{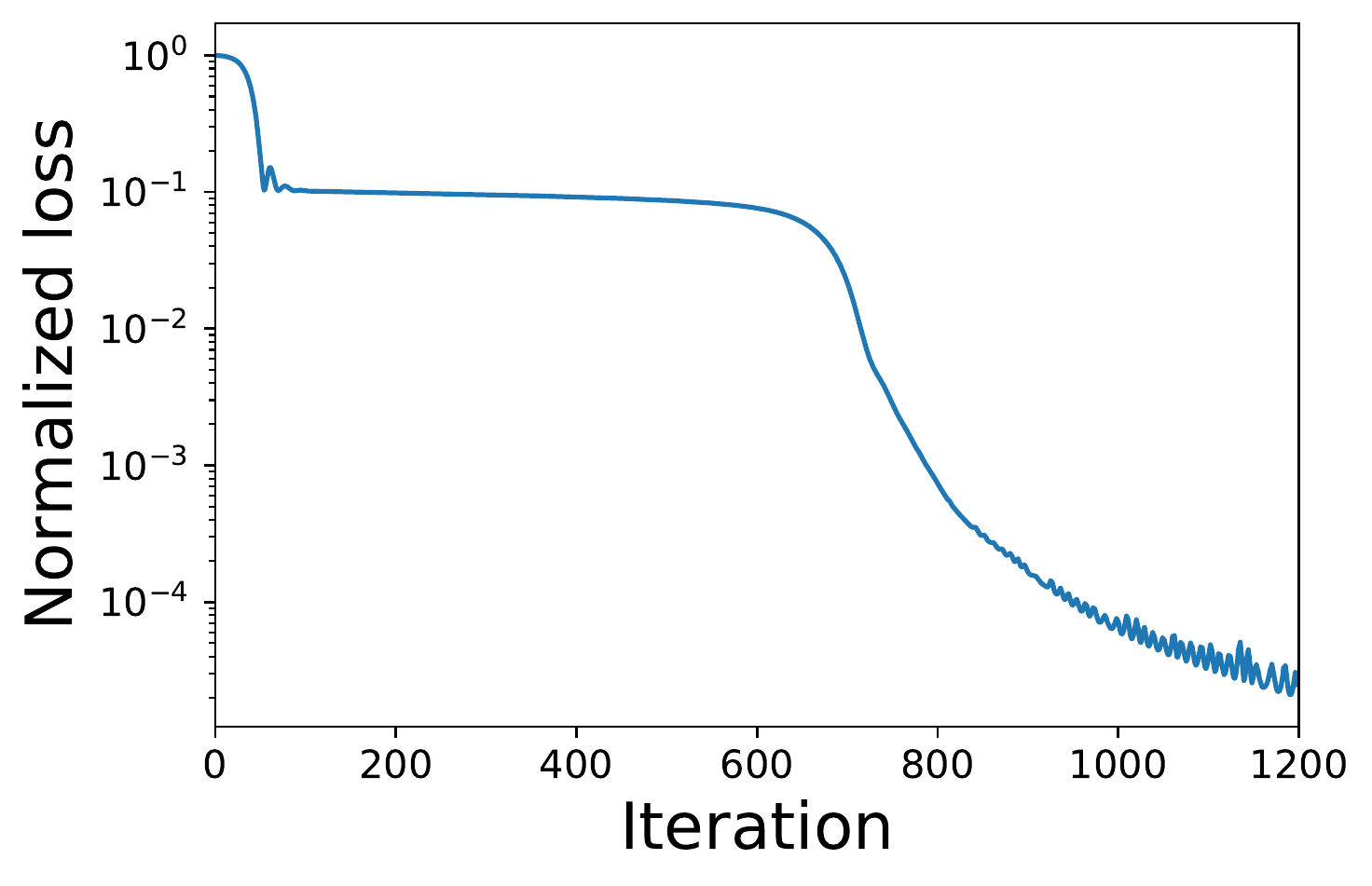}}
\caption{Case Study 2: Stiff Single ODE example. a)  Exact solution vs. Reduced-PINN predictions, b) Convergence history }
\label{fig56}
\end{figure}

The complex behavior in stiff ODE is induced by the presence of steep gradients which often forms a significant computational barrier in computational efforts.
Stiff ODEs can be mainly identified by different time-scales of the associated species. 
In long-time domain problems, a decomposition of the temporal domain into sub-intervals can minimize the carryover of the error during the chemical reactions between the species. Accordingly, the following initial value problem is used to illustrate capabilities of Reduced-PINN
 \begin{equation}
\label{eq32}
u^{(1)}(x)-\lambda u(x) = e^{-x}, (0\le{x}\le0.05)
\end{equation}
 
\begin{equation}
\label{eq33}
\ u(0) = \mu^{<0>}
\end{equation}
The exact solution of the above IVP is written as 
\begin{equation}
\label{eq34}
u_{exact}(x) =\left( \mu^{<0>} + \frac{1}{1 + \lambda}\right)e^{\lambda x}-\frac{e^{-x}}{1 + \lambda}
\end{equation}
and its integral form is
\begin{equation}
\label{eq35}
v(x)-\lambda\int_{0}^{x}v(t)dt = e^{-x} + \lambda \mu^{<0>}
\end{equation}
 also, after training and obtaining $v(x)$, the Reduced-PINN solution to our IVP can be derived from
\begin{equation}
\label{eq36}
u(x) = \int_{0}^{x} v(t)dt + \mu^{<0>}
\end{equation}
If we take $\lambda$ to be a large value in magnitude, then the exact solution includes both fast and slow components. To address the issue, we set $\lambda = -50$ and $\mu^{<0>}=2$. Our method works quite well for this stiff scalar ODE and numerical results agree with the exact solution as shown in Fig.\ref{fig56}a. We also provide the training history of our method in Fig.\ref{fig56}b.

\subsection{Case Study 3: Stiff ODE System} 
Next, we benchmarked Reduced-PINN against the following system of stiff linear ODEs that were used to benchmark stiff-ODE solvers by \cite{ref20} 
\begin{align}
\label{eq37}
u_{1}^{(1)}(x) &= (\frac{\lambda_{1}+\lambda_{2}}{2})u_{1}(x) + (\frac{\lambda_{1}-\lambda_{2}}{2})u_{2}(x) \nonumber \\
u_{2}^{(1)}(x) &= (\frac{\lambda_{1}-\lambda_{2}}{2})u_{1}(x) + (\frac{\lambda_{1}+\lambda_{2}}{2})u_{2}(x)
\end{align}
with initial conditions $u_1(0) = \mu^{<0>}_1$ and $u_2(0) = \mu^{<0>}_2$. The solution of the given system of ODEs is written as
\begin{equation}
\label{eq39}
u_{1}(x) = (\frac{\mu^{<0>}_1+\mu^{<0>}_2}{2})e^{\lambda_1x} + (\frac{\mu^{<0>}_1-\mu^{<0>}_2}{2})e^{\lambda_2x}
\end{equation}
\begin{equation}
\label{eq40}
u_{2}(x) = (\frac{\mu^{<0>}_1+\mu^{<0>}_2}{2})e^{\lambda_1x} - (\frac{\mu^{<0>}_1-\mu^{<0>}_2}{2})e^{\lambda_2x}
\end{equation}
To put the equations into integral form, let $u_{1}^{(1)}(x) = v_1(x)$ and $u_{2}^{(1)}(x) = v_2(x)$, then we have
\begin{equation}
\label{eq41}
u_1(x) = \int_{0}^{x}v_1(t) dt+\mu^{<0>}_1
\end{equation}
\begin{equation}
\label{eq42}
u_2(x) = \int_{0}^{x}v_2(t) dt+\mu^{<0>}_2
\end{equation}
Substituting the above expressions into our system of equations yields
\begin{equation}
\label{eq43}
v_1(x) - (\frac{\lambda_{1}+\lambda_{2}}{2})(\int_{0}^{x}v_1(t) dt+\mu^{<0>}_1)-(\frac{\lambda_{1}-\lambda_{2}}{2})(\int_{0}^{x}v_2(t) dt+\mu^{<0>}_2)=0
\end{equation}
\begin{equation}
\label{eq44}
v_2(x) - (\frac{\lambda_{1}-\lambda_{2}}{2})(\int_{0}^{x}v_1(t) dt+\mu^{<0>}_1)-(\frac{\lambda_{1}+\lambda_{2}}{2})(\int_{0}^{x}v_2(t) dt+\mu^{<0>}_2)=0
\end{equation}

We have considered the following parameters for the given system: $\lambda_1 = -20$, $\lambda_2 = -2$, $\mu^{<0>}_1 = 2$, $\mu^{<0>}_2 = 0$ and $(0\le{x}\le1.0)$. To solve this problem, we have used five sequential Reduced-PINNs with identical parameters. The time interval considered for each of them has the same length. 

\begin{figure}
\centering
\footnotesize{a)}{\includegraphics[width = 0.46\linewidth]{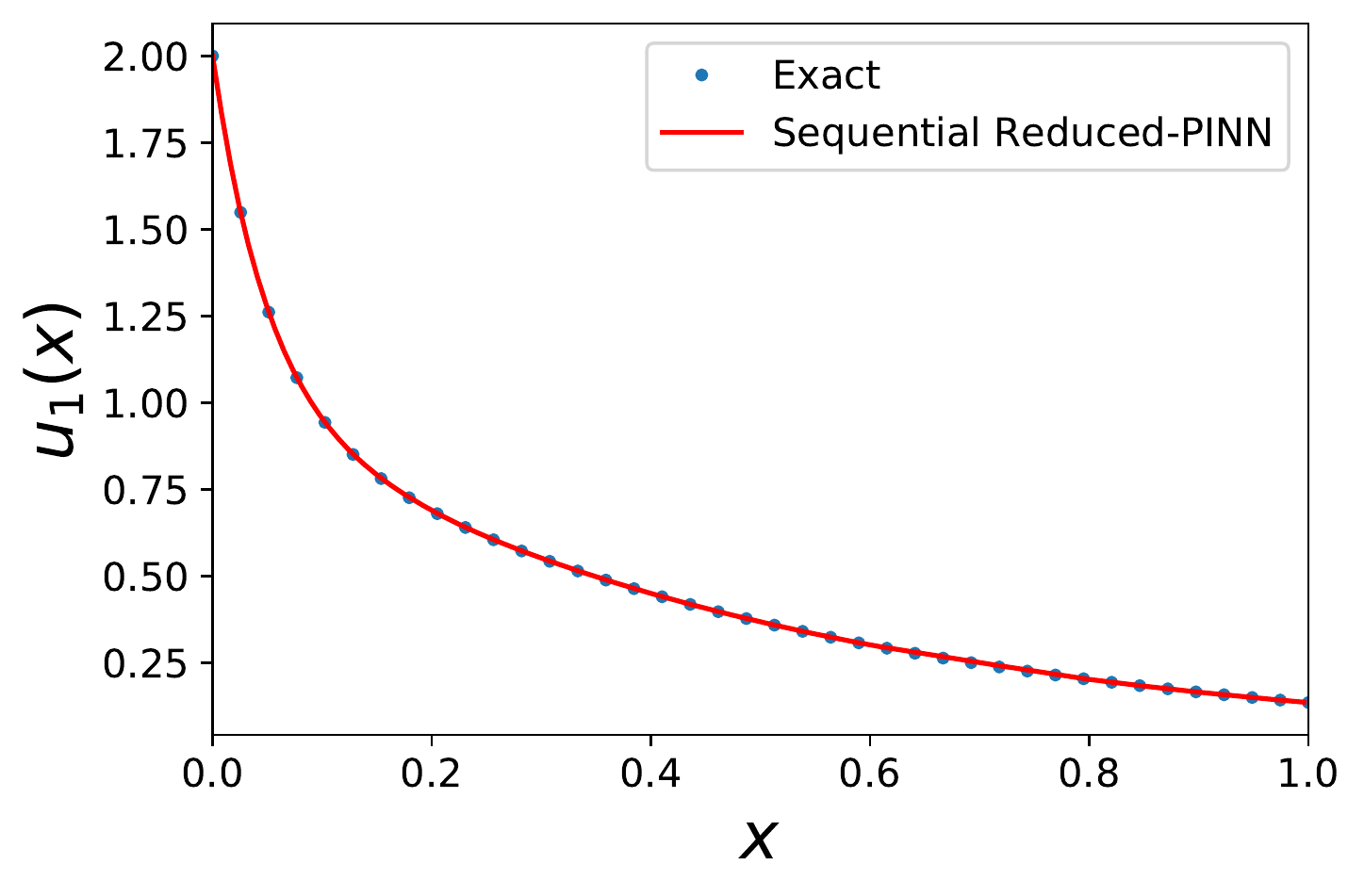}}
\footnotesize{b)}{\includegraphics[width = 0.46\linewidth]{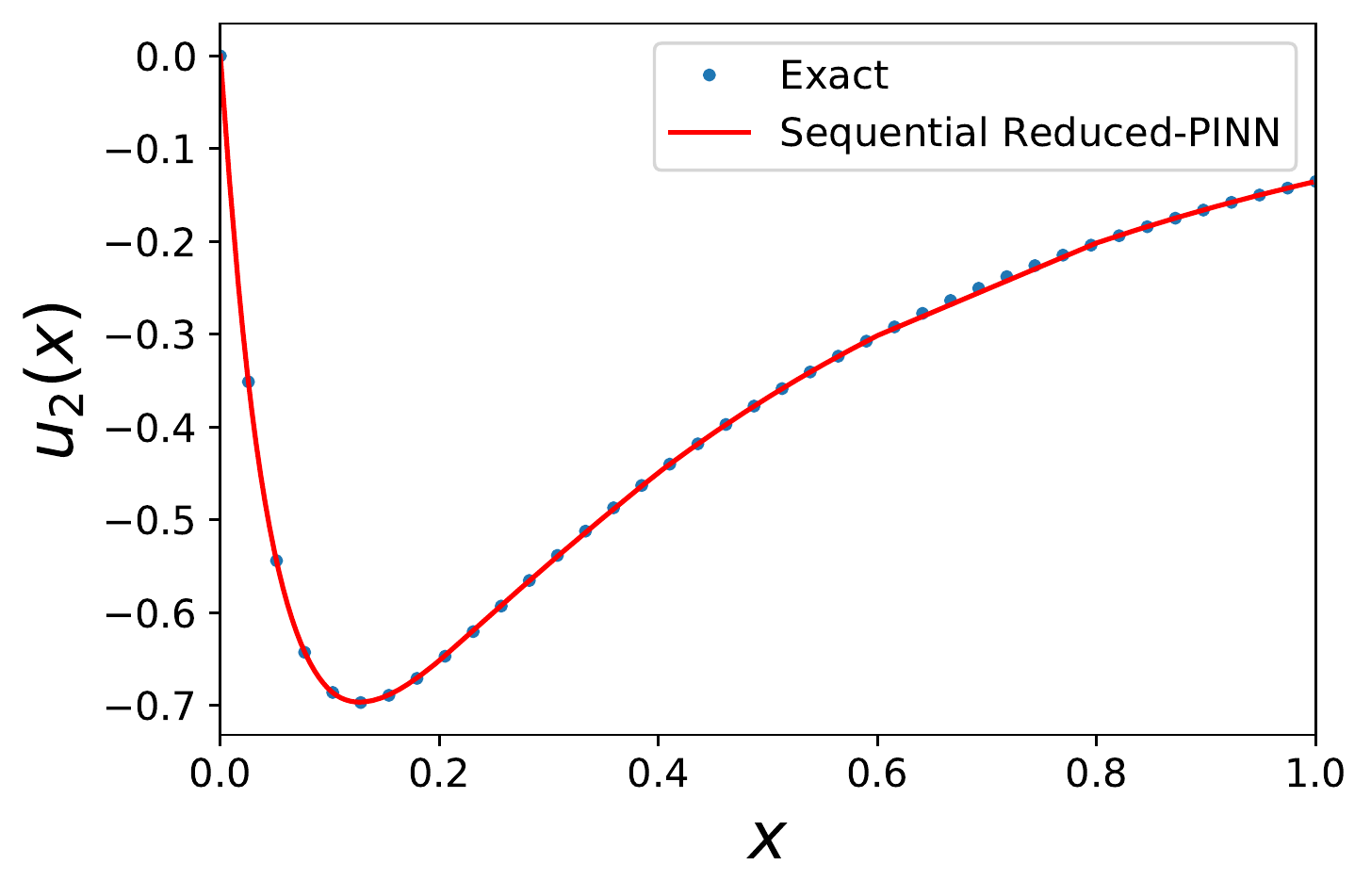}}
\caption{Case Study 3: Stiff ODE System (a) Exact solution vs. the proposed method for the first state variable (b) Exact solution vs. the proposed method for the second state variable}
\label{fig7}
\end{figure}

 Since there are two unknown functions, we have chosen a deep neural network for each function. The left hand sides of Eqs. (\ref{eq43}) and (\ref{eq44}) are the residuals associated with the first and second ODEs, respectively. Therefore, the total loss function consists of two parts. Fig.\ref{fig7} shows numerical results vs. exact solutions for $u_1(x)$ and $u_2(x)$. As can be seen from this figure, we have an excellent agreement between Reduced-PINN and exact solution for this stiff system of ODEs.
 
\subsection{Case Study 4: ROBER Problem}
\begin{figure}
\centering
\footnotesize{a)}{\includegraphics[width = 0.45\linewidth]{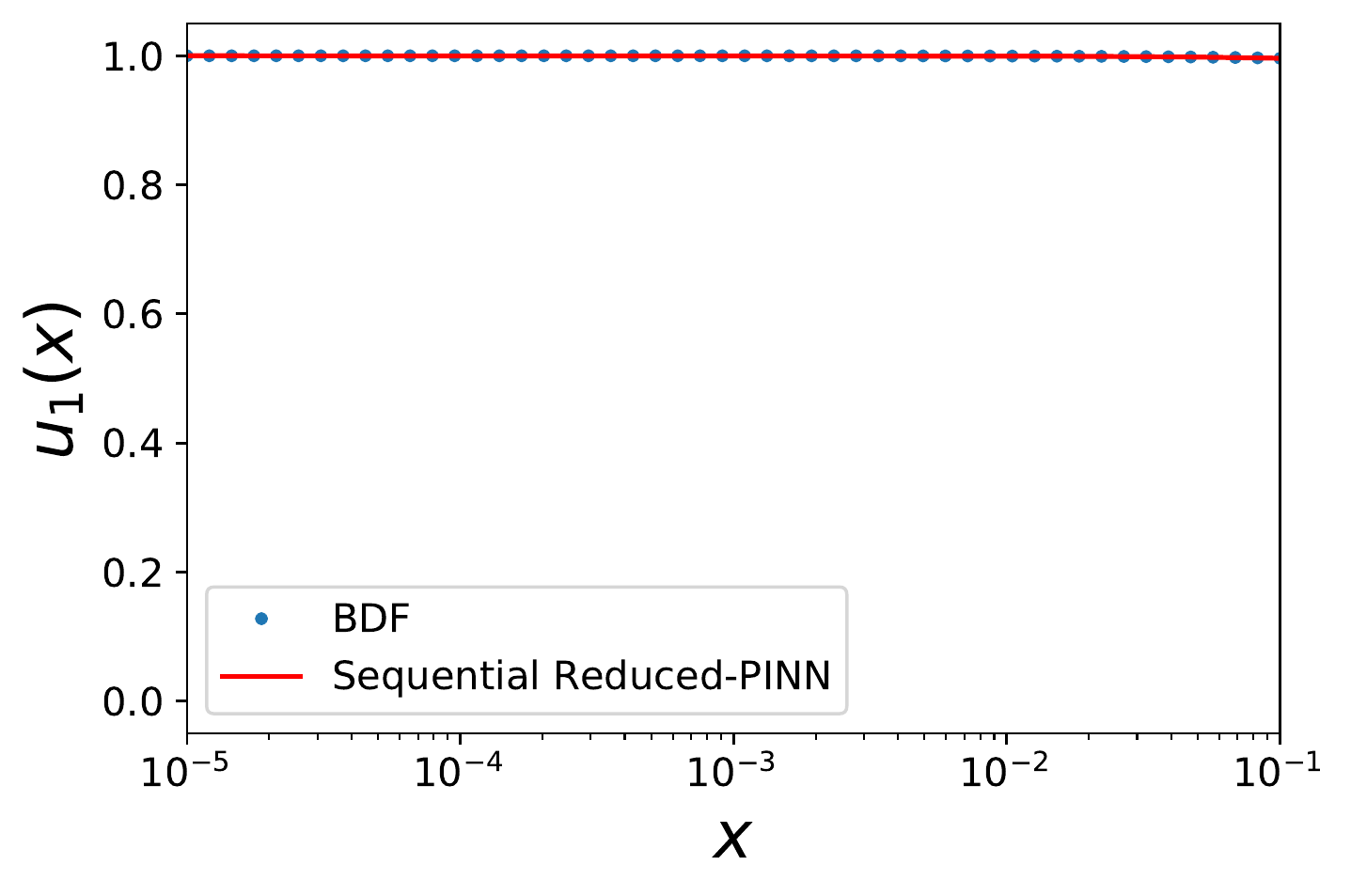}} \hfill
\footnotesize{b)}{\includegraphics[width = 0.45\linewidth]{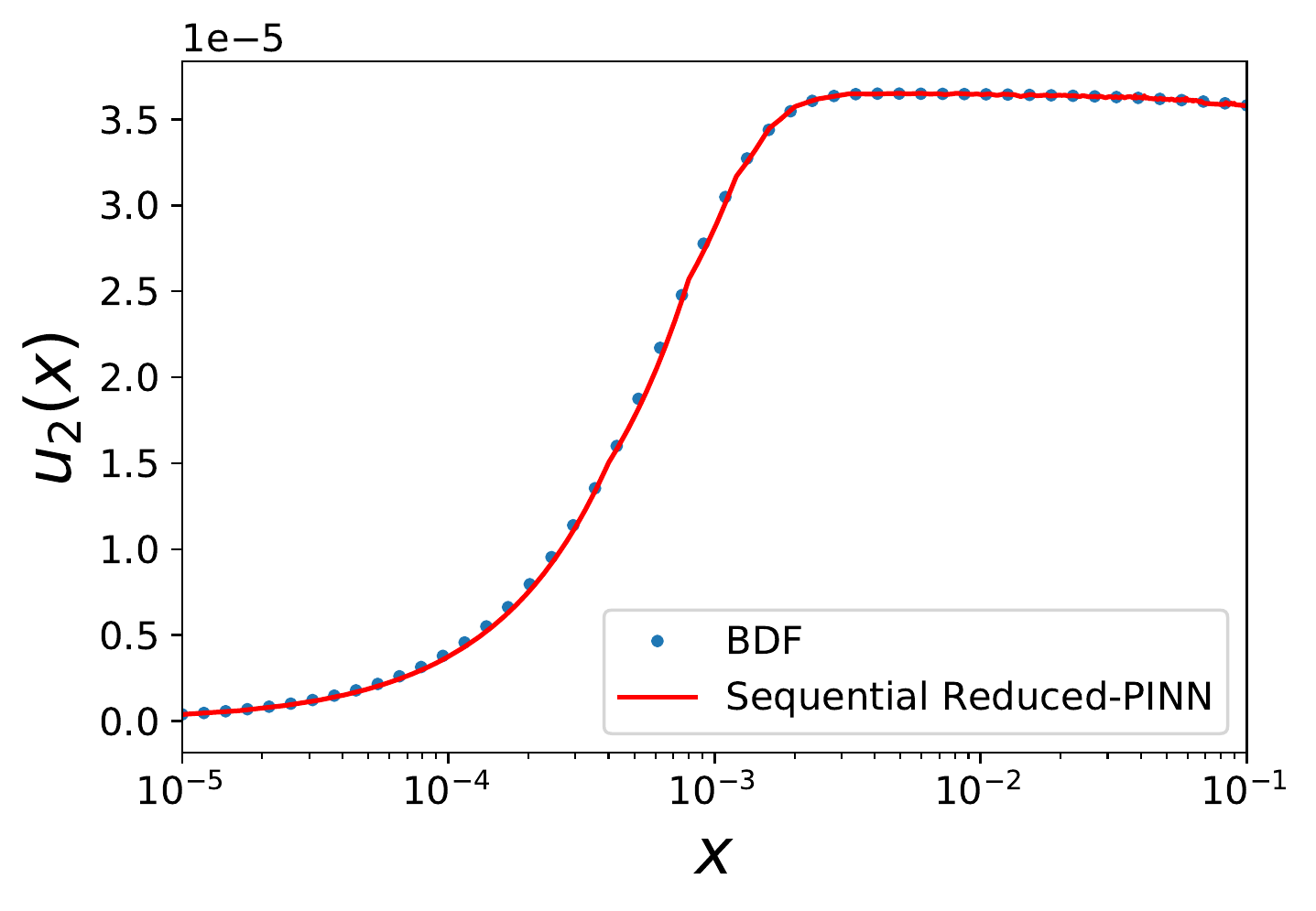}}\\
\center
\footnotesize{c)}{\includegraphics[width = 0.45\linewidth]{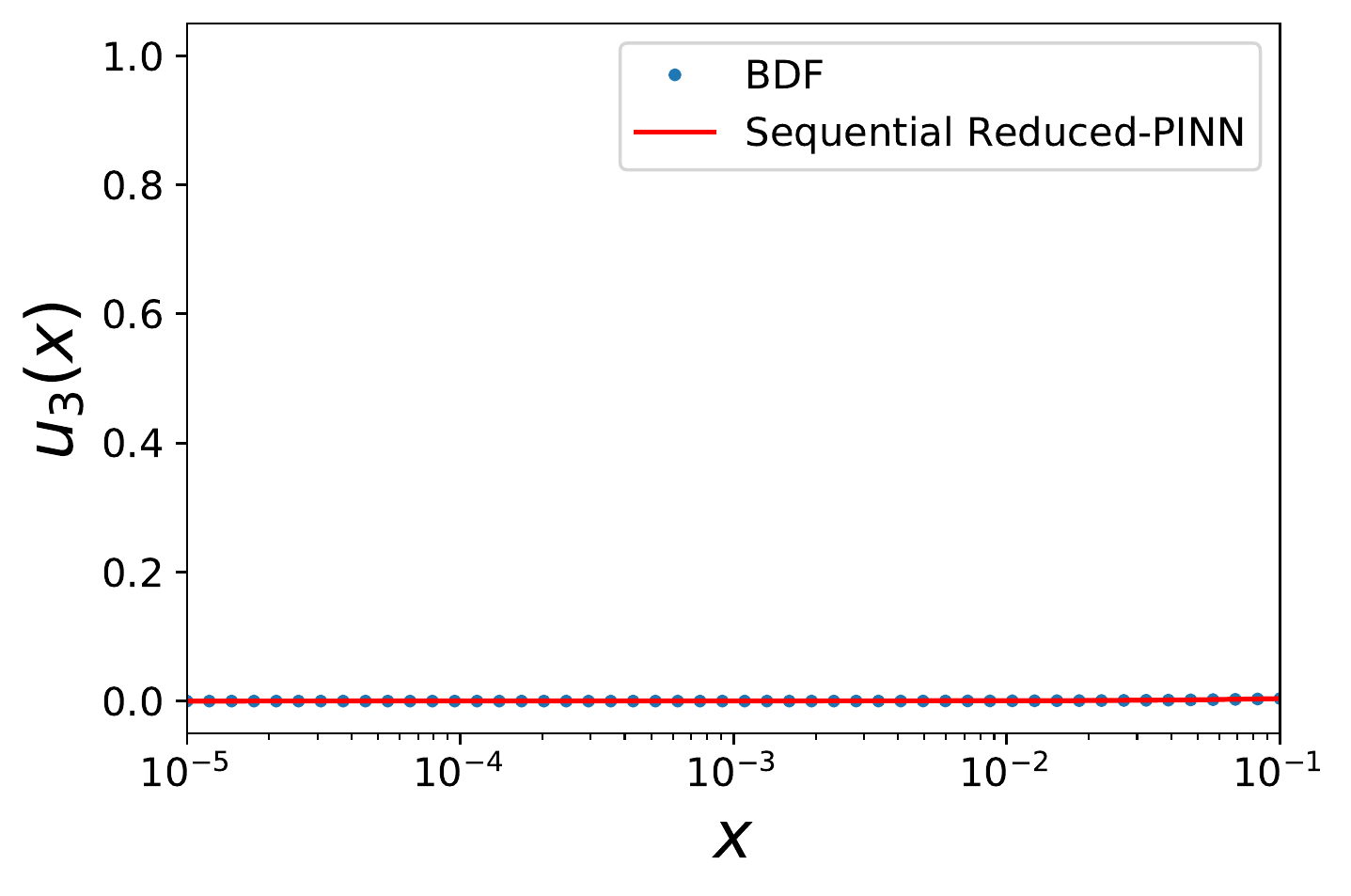}}
\caption{Case Study 4: ROBER Problem; Comparison between the BDF method and sequential Reduced-PINN for ROBER problem (a) Evolution of the first species concentration (b) Evolution of the second species concentration (c) Evolution of the third species concentration} 
\label{fig8}
\end{figure}

 Our last example is Robertson's system of equations \cite{ref21} referred to as the ROBER problem first by Wanner and Hairer \cite{ref22}
It describes the following auto-catalytic reaction kinetics

\begin{align}
	A &\underset{k_1}{ \rightarrow }B\nonumber \\
	B+B &\underset{k_2}{ \rightarrow }C+B\nonumber \\
		B+C &\underset{k_2}{ \rightarrow }A+C,
\end{align}
where $A, B$, and $C$ are the chemical species concentrations, and $k_1, k_2$, and $k_3$ the reaction rate constants. The ROBER problem can be written using three nonlinear ODEs as follows
 \begin{align}
 \label{eq45}
 u_{1}^{(1)}(x) &= -k_{1}u_{1} + k_{3} u_{2} u_{3} \qquad u_1(0) = 1 \nonumber \\
 u_{2}^{(1)}(x) &= k_{1}u_{1} - k_{2} u_{2}^2 - k_{3} u_{2} u_{3} \qquad u_2(0) = 0 \nonumber\\
 u_{3}^{(1)}(x) &= k_{2} u_{2}^2 \qquad u_3(0) = 0. 
 \end{align}
 Concentrations of species in the above equation are represented by $u_i(x)$ and the reaction constants are $k_1 = 0.04$, $k_2= 3 \times10^7$, and $k_3= 10^4$ which cover a time-span in the range of 9 orders. 
	Such a wide gap in time-span of the species leads to the strong stiffness of the problem.
Here, we have considered 250 sequential Reduced-PINNs with identical parameters to solve the problem for the time duration $[10^{-5}, 10^{-1}]$. Also, backward differentiation formula (BDF) is utilized as an implicit integrator for the purpose of comparison. The results obtained from Reduced-PINN for species concentrations are plotted in Fig.\ref{fig8}. As shown in this figure, we have an excellent agreement between Reduced-PINN and BDF method results. 
Our results show that the Reduced-PINN method can be reliably used to solve the ROBER problem, which classical PINN fails to do, even if
artifacts to reduce the stiffness of the ODE system are hired, as in Stiff-PINNs \cite{ref13}.

 \section{Outlook} 
 Future works will focus on developing a robust and general Reduced-PINN framework for stiff chemical kinetic simulations that can handle multiple volatile species with extremely short time scales. Our goal is to demonstrate the relevance of the Reduced-PINN framework for local residuals (e.g., chemically reacting flows) and non-local residuals (e.g., PDEs with spatial derivatives) compared to  PINNs and traditional DNNs. 
\section{Conclusions} 
 
We proposed a new neural network engine based on PINN for solving stiff initial value problems using integral-based physics-informed neural networks. This study is motivated by simply using the residual of the PDE as in Physics-Informed Neural Networks (PINNs) is not optimal for the simulation of multi-species chemical reactions. Our engine, referred to as Reduced-PINN, was developed based on the concept of converting ODEs into a weak-form integral equation. This approach ensures we combine loss functions and boundary conditions into one loss equation. Next, we show how to build the total loss function for a general system of ODEs. We benchmarked Reduced-PINN prediction against four systems of varying complexity to validate its capabilities.
Using Reduced-PINN sequentially is more suitable for solving initial value problems for long-term problems. Next, we used Reduced-PINNs to represent stiff ODE systems, for which the regular-PINN failed to predict the evolution of the systems. By imposing reduced weak form integral on loss function of involved species in the kinetic systems and reducing the stiffness, the Stiff-PINN well captured the dynamic responses of the reference ODE systems. Our numerical benchmarking shows the Reduced-PINN efficacy and accuracy in solving stiff ODE systems at a much lower cost.

 \bibliographystyle{ieeetr}
 \bibliography{paper}

\begin{thebibliography}{10}

\bibitem{ref1}
A.~Iserles, {\em A first course in the numerical analysis of differential
  equations}.
\newblock No.~44, Cambridge university press, 2009.

\bibitem{ref2}
M.~Raissi, P.~Perdikaris, and G.~E. Karniadakis, ``Physics-informed neural
  networks: A deep learning framework for solving forward and inverse problems
  involving nonlinear partial differential equations,'' {\em Journal of
  Computational physics}, vol.~378, pp.~686--707, 2019.

\bibitem{ref3}
E.~Haghighat, M.~Raissi, A.~Moure, H.~Gomez, and R.~Juanes, ``A
  physics-informed deep learning framework for inversion and surrogate modeling
  in solid mechanics,'' {\em Computer Methods in Applied Mechanics and
  Engineering}, vol.~379, p.~113741, 2021.

\bibitem{ref4}
M.~Vahab, E.~Haghighat, M.~Khaleghi, and N.~Khalili, ``A physics-informed
  neural network approach to solution and identification of biharmonic
  equations of elasticity,'' {\em Journal of Engineering Mechanics}, vol.~148,
  no.~2, p.~04021154, 2022.

\bibitem{ref5}
S.~A. Niaki, E.~Haghighat, T.~Campbell, A.~Poursartip, and R.~Vaziri,
  ``Physics-informed neural network for modelling the thermochemical curing
  process of composite-tool systems during manufacture,'' {\em Computer Methods
  in Applied Mechanics and Engineering}, vol.~384, p.~113959, 2021.

\bibitem{ref6}
E.~Samaniego, C.~Anitescu, S.~Goswami, V.~M. Nguyen-Thanh, H.~Guo, K.~Hamdia,
  X.~Zhuang, and T.~Rabczuk, ``An energy approach to the solution of partial
  differential equations in computational mechanics via machine learning:
  Concepts, implementation and applications,'' {\em Computer Methods in Applied
  Mechanics and Engineering}, vol.~362, p.~112790, 2020.

\bibitem{ref7}
R.~Arora, P.~Kakkar, B.~Dey, and A.~Chakraborty, ``Physics-informed neural
  networks for modeling rate-and temperature-dependent plasticity,'' {\em arXiv
  preprint arXiv:2201.08363}, 2022.

\bibitem{ref8}
Z.~Mao, A.~D. Jagtap, and G.~E. Karniadakis, ``Physics-informed neural networks
  for high-speed flows,'' {\em Computer Methods in Applied Mechanics and
  Engineering}, vol.~360, p.~112789, 2020.

\bibitem{ref9}
M.~M. Almajid and M.~O. Abu-Al-Saud, ``Prediction of porous media fluid flow
  using physics informed neural networks,'' {\em Journal of Petroleum Science
  and Engineering}, vol.~208, p.~109205, 2022.

\bibitem{ref10}
A.~D. Jagtap, E.~Kharazmi, and G.~E. Karniadakis, ``Conservative
  physics-informed neural networks on discrete domains for conservation laws:
  Applications to forward and inverse problems,'' {\em Computer Methods in
  Applied Mechanics and Engineering}, vol.~365, p.~113028, 2020.

\bibitem{ref11}
E.~Kharazmi, Z.~Zhang, and G.~E. Karniadakis, ``Variational physics-informed
  neural networks for solving partial differential equations,'' {\em arXiv
  preprint arXiv:1912.00873}, 2019.

\bibitem{ref12}
E.~Kharazmi, Z.~Zhang, and G.~E. Karniadakis, ``hp-vpinns: Variational
  physics-informed neural networks with domain decomposition,'' {\em Computer
  Methods in Applied Mechanics and Engineering}, vol.~374, p.~113547, 2021.

\bibitem{ref13}
W.~Ji, W.~Qiu, Z.~Shi, S.~Pan, and S.~Deng, ``Stiff-pinn: Physics-informed
  neural network for stiff chemical kinetics,'' {\em The Journal of Physical
  Chemistry A}, vol.~125, no.~36, pp.~8098--8106, 2021.

\bibitem{ref14}
M.~De~Florio, E.~Schiassi, and R.~Furfaro, ``Physics-informed neural networks
  and functional interpolation for stiff chemical kinetics,'' {\em Chaos: An
  Interdisciplinary Journal of Nonlinear Science}, vol.~32, no.~6, p.~063107,
  2022.

\bibitem{ref15}
S.~Wang, Y.~Teng, and P.~Perdikaris, ``Understanding and mitigating gradient
  flow pathologies in physics-informed neural networks,'' {\em SIAM Journal on
  Scientific Computing}, vol.~43, no.~5, pp.~A3055--A3081, 2021.

\bibitem{ref16}
J.~D. Lambert, {\em Numerical methods for ordinary differential systems: the
  initial value problem}.
\newblock John Wiley \& Sons, Inc., 1991.

\bibitem{ref17}
L.~G. Chambers, {\em Integral equations: a short course}.
\newblock International Textbook Company, 1976.

\bibitem{ref18}
A.~G. Baydin, B.~A. Pearlmutter, A.~A. Radul, and J.~M. Siskind, ``Automatic
  differentiation in machine learning: a survey,'' {\em Journal of Marchine
  Learning Research}, vol.~18, pp.~1--43, 2018.

\bibitem{ref19}
M.~Abadi, P.~Barham, J.~Chen, Z.~Chen, A.~Davis, J.~Dean, M.~Devin,
  S.~Ghemawat, G.~Irving, M.~Isard, {\em et~al.}, ``$\{$TensorFlow$\}$: A
  system for $\{$Large-Scale$\}$ machine learning,'' in {\em 12th USENIX
  symposium on operating systems design and implementation (OSDI 16)},
  pp.~265--283, 2016.

\bibitem{ref20}
R.~Bulirsch, J.~Stoer, and J.~Stoer, {\em Introduction to numerical analysis},
  vol.~3.
\newblock Springer, 2002.

\bibitem{ref21}
H.~Robertson, ``The solution of a set of reaction rate equations,'' {\em
  Numerical analysis: an introduction}, vol.~178182, 1966.

\bibitem{ref22}
G.~Wanner and E.~Hairer, {\em Solving ordinary differential equations II},
  vol.~375.
\newblock Springer Berlin Heidelberg New York, 1996.

\end{thebibliography}
\end{document}